\documentclass[graybox]{svmult}

\usepackage{type1cm}        
\usepackage{makeidx}         
\usepackage{graphicx}        
\usepackage{multicol}        
\usepackage[bottom]{footmisc}
\usepackage{subfig}

\usepackage{newtxtext}       %
\usepackage{newtxmath}       

\makeindex             

\begin{document}

\graphicspath{{Zollhoefer/}}

\title{Commodity RGB-D Sensors: Data Acquisition}

\author{Michael Zollh\"ofer}

\institute{Michael Zollh\"ofer \at Stanford University, 353 Serra Mall, Stanford, CA 94305, USA, \email{zollhoefer@cs.stanford.edu}}

\maketitle
\label{Zollhoefer}


\abstract
{
	Over the past ten years we have seen a democratization of range sensing technology.
	While previously range sensors have been highly expensive and only accessible to a few domain experts, such sensors are	nowadays ubiquitous and can even be found in the latest generation of mobile devices, e.g., current smartphones.
	This democratization of range sensing technology was started with the release of the Microsoft Kinect, and since then many different commodity range sensors followed its lead, such as the Primesense Carmine, Asus Xtion Pro, and the Structure Sensor from Occipital.
	The availability of cheap range sensing technology led to a big leap in research, especially in the context of more powerful static and dynamic reconstruction techniques, starting from 3D scanning applications, such as KinectFusion, to highly accurate face and body tracking approaches.
	In this chapter, we have a	detailed look into the different types of existing range sensors.
	We discuss the two fundamental types of commodity range sensing techniques in detail, namely passive and active sensing, and we explore the principles these technologies are based on.
	%
	%
	Our focus is on modern active commodity range sensors based on time-of-flight and structured light.
	We conclude by discussing the noise characteristics, working ranges, and types of errors made by the different sensing modalities.
}

\index{Depth Sensor}
\index{Range Sensor}
\index{RGB-D Sensor}

\section{Introduction} \label{MZsec:1}
\begin{figure}[t]
	\centering
	\sidecaption
	\subfloat[Color]{{\includegraphics[width=3.6cm]{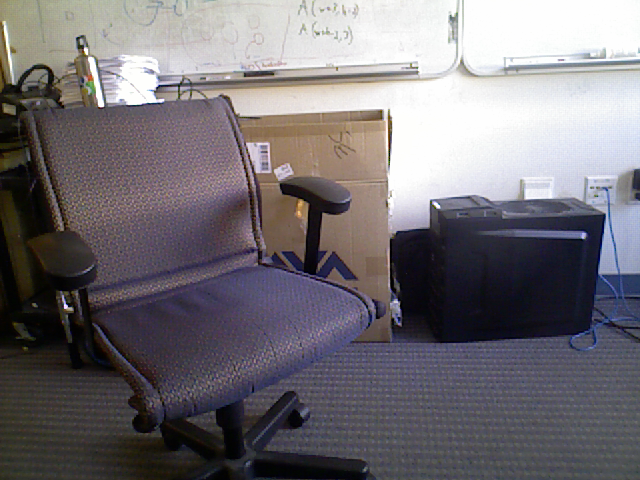} }}%
	\quad
	\subfloat[Depth]{{\includegraphics[width=3.6cm]{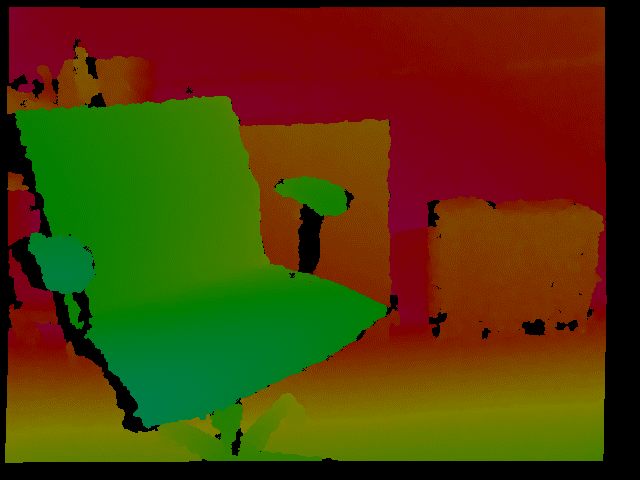} }}%
	\quad
	\subfloat[Phong]{{\includegraphics[width=3.6cm]{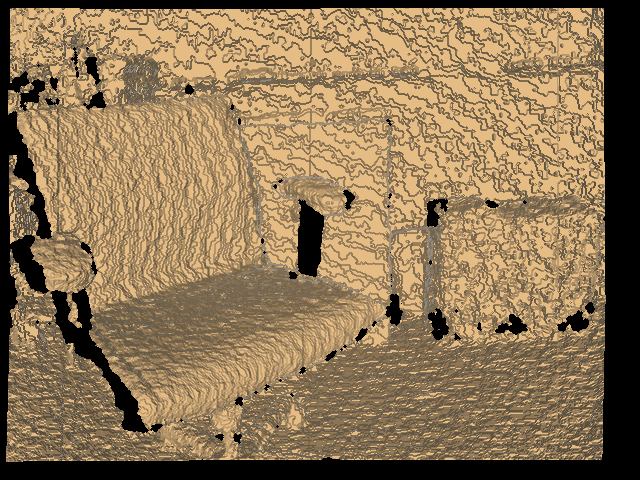} }}%
	\caption{An RGB-D camera jointly captures color (a) and depth (b) images. The depth image encodes the distance to the scene on a per-pixel basis. Green color means that this part of the scene is close to the camera and red means that it is far away. The phong shaded image (c) is an alternative visualization of the 3D geometry.
	}
	\label{fig:rgbd}
\end{figure}
Modern conventional color cameras are ubiquitous in our society and enable us to capture precious memories in a persistent and digital manner.
These recordings are represented as millions of three channel pixels that encode the amount of red, green, and blue light that reached the sensor at a corresponding sensor location and time.
Unfortunately, color images are an inherently flat 2D representation, since most of the 3D scene information is lost during the process of image formation.

Over the past ten years we have seen a democratization of a new class of cameras that enables the dense measurement of the 3D geometry of the observed scene, thus overcoming the mentioned limitation of conventional color cameras.
These so called \textit{range} or \textit{depth sensors} perform a dense per-pixel measurement of scene depth, i.e., the distance to the observed points in the scene.
These measured depth values are normally exposed to the user in the form of a \textit{depth image}, which is a $2.5$-dimensional representation of the visible parts of the scene.
An \textit{RGB-D sensor} is the combination of a conventional color camera (RGB) with such a depth sensor (D).
It enables the joint capture of scene appearance and scene geometry at real-time frame rates based on a stream of color $\mathcal{C}$ and depth images $\mathcal{D}$.
Figure \ref{fig:rgbd} shows an example of such a color (a) and depth image pair (b).
The phong shaded image (c) is an alternative visualization of the captured 3D geometry that better illustrates the accuracy of the obtained depth measurements.
Current RGB-D sensors provide a live stream of color and depth at over 30\,Hz.

Starting with the \textit{Microsoft Kinect}, over the past 10 years a large number of commodity RGB-D sensors have been developed, such as the \textit{Primesense Carmine}, \textit{Asus Xtion Pro}, \textit{Creative Senz3D}, \textit{Microsoft Kinect One}, \textit{Intel Realsense} and the \textit{Structure Sensor}.
While previous range sensors \cite{proto0,proto1,proto2} where highly expensive and only accessible to a few domain experts, range sensors are nowadays ubiquitous and can even be found in the latest generation of mobile devices.
Current sensors have a small form factor, are affordable, and accessible for everyday use to a broad audience.
The availability of cheap range sensing technology led to a big leap in research \cite{overview0}, especially in the context of more powerful static and dynamic reconstruction techniques, starting from 3D scanning applications, such as KinectFusion, to highly accurate face and body tracking approaches.
One very recent example is the current Apple iPhone X that employs the range data captured by an off-the-shelf depth sensor as part of its face identification system.

In the following, we review the technical foundations of such camera systems.
We will start by reviewing the \textit{Pinhole Camera} model and perspective projections.
Afterwards, we will introduce the ideas behind both \textit{passive} as well as \textit{active} depth sensing approaches and explain their fundamental working principles.
More specifically, we will discuss how commodity RGB-D sensors based on \textit{Stereo Vision} (SV), \textit{Structured Light} (SL) and \textit{Time-of-Flight} (ToF) technology work.
We conclude by comparing the different depth sensing modalities and discussing their advantages and disadvantages.

\index{Pinhole Camera}
\index{Perspective Camera}
\index{Intrinsic Camera Parameters}
\index{Focal Length}
\index{Principal Point}
\section{Projective Camera Geometry} \label{MZsec:2}
%
%
We start by reviewing the \textit{Pinhole Camera} model, which is a simplified version of the projective geometry of real world cameras, since it is a basic building block for many types of depth sensors.
\begin{figure}[t]
	\centering
	\sidecaption
	\subfloat[3D View]{{\includegraphics[width=5.5cm]{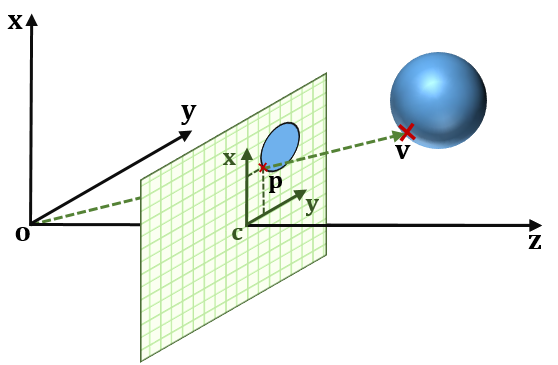} }}%
	\quad
	\subfloat[Side View]{{\includegraphics[width=5.5cm]{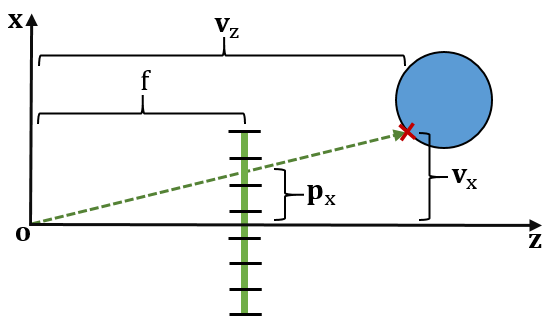} }}%
	\caption{Perspective camera geometry. The image sensor is shown in green. The \textit{Pinhole Camera} model describes how a point $\mathbf{v} \in \mathbb{R}^3$ is mapped to a location $\mathbf{p} \in \mathbb{R}^2$ on the sensor. The $\mathbf{z}$-axis is the cameras viewing direction and the $\mathbf{x}$-axis is the up-vector. The perspective projection is defined by the camera's focal length $f$ and the principal point $\mathbf{c}$. The focal length $f$ is the distance between the sensor plane and the origin $\mathbf{o}$ of the camera coordinate system.}
	\label{fig:camera}
\end{figure}
An illustration of the perspective projection defined by the \textit{Pinhole Camera} model can be found in Figure~\ref{fig:camera}.
A 3D point $\mathbf{v} = (\mathbf{v}_x, \mathbf{v}_y, \mathbf{v}_z)^T \in \mathbb{R}^3$ in camera space is mapped to the sensor plane (green) based on a perspective projection \cite{projection0}.
The resulting point $\mathbf{p} = (\mathbf{p}_x, \mathbf{p}_y)^T \in \mathbb{R}^2$ on the sensor depends on the intrinsic properties of the camera, i.e., its focal length $f$ and the principal point $\mathbf{c} = (\mathbf{c}_x, \mathbf{c}_y)^T$. 
Let us first assume that the principal point is at the center of the sensor plane, i.e., $\mathbf{c}=(0, 0)^T$.
In the following, we show how to compute the 2D position $\mathbf{p}$ on the image plane given a 3D point $\mathbf{x}$ and the intrinsic camera parameters.
By applying the geometric rule of equal triangles, the following relation can be obtained, see also Figure \ref{fig:camera} for an illustration:
\begin{eqnarray}
	\frac{\mathbf{p}_x}{f} &=& \frac{\mathbf{v}_x}{\mathbf{v}_z} \enspace{.}
	\label{eq:01}
\end{eqnarray}
With the same reasoning, a similar relation also holds for the $y$-component.
Reordering and solving for $\mathbf{p}$ leads to the fundamental equations of perspective projection that describe how a 3D point $\mathbf{v}$ is projected to the sensor plane:
\begin{eqnarray}
	\mathbf{p}_x &=& \frac{f \cdot \mathbf{v}_x}{\mathbf{v}_z}\enspace{,}\\
	\mathbf{p}_y &=& \frac{f \cdot \mathbf{v}_y}{\mathbf{v}_z}\enspace{.}
	\label{eq:02}
\end{eqnarray}
The same mapping can be more concisely represented in matrix-vector notation by using homogeneous coordinates.
Let $\mathbf{K}$ be the intrinsic camera matrix:
\begin{equation}
	\mathbf{K} =
	\begin{bmatrix}
		f & s & \mathbf{c}_x \\
		0 & f & \mathbf{c}_y \\
		0 & 0 & 1   \\
	\end{bmatrix} \enspace{.}
\end{equation}
Here, $s$ is an additional skew parameter \cite{overview1} and $\mathbf{c}$ specifies the principal point, which we assumed to be zero so far.
Given the definition of $\mathbf{K}$, the perspective projection can be represented as $\hat{\mathbf{p}} = \mathbf{K}{\mathbf{v}}$, where $\hat{\mathbf{p}} \in \mathbb{R}^3$ are the homogeneous coordinates of the 2D point $\mathbf{p}$.
%
%
The intrinsic camera parameters can be obtained based on camera calibration routines \cite{calibration0,calibration1}.
The \textit{Pinhole Camera} model is one of the basic building blocks of range sensing approaches.
It makes a few simplifying assumptions, such as that the lens is perfect, i.e., that there are no lens distortions.
Lens distortion \cite{distortion0} can be tackled in a preprocessing step by calibrating the camera.

\index{Range Sensing!Passive Sensing}
\index{Triangulation}
\section{Passive Range Sensing} \label{MZsec:3}
Similar to human 3D vision, passive range sensing is implemented based on the input of two or multiple \cite{multi0} conventional monochrome or color cameras.
\begin{figure}[t]
	\centering
	\includegraphics[scale=.7]{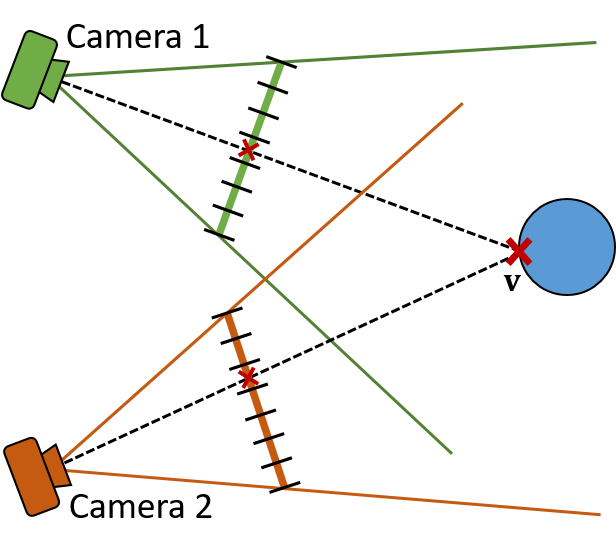}
	\caption{Stereo reconstruction. Similar to human vision, stereo approaches employ two cameras to obtain observations of the scene from two slightly different viewpoints. In the first step of stereo reconstruction, corresponding points in both images are computed, i.e., pixels of the images that observe the same 3D point in the scene. Based on these matches, the 3D position can be found via triangulation, i.e., by intersecting two rays cast through the detected point correspondences.}
	\label{fig:stereo}
\end{figure}
Here, the term `passive' refers to the fact that passive sensors do not modify the scene to obtain the scene depth.
The special case of obtaining depth measurements based on only two cameras \cite{stereo0} is known as \textit{stereo} or \textit{binocular reconstruction}.
These systems are quite cheap and have a low power consumption, since they are based on two normal color cameras.
The basic setup of such a stereo camera system is illustrated in Figure~\ref{fig:stereo}.

Scene depth can be estimated based on a computational process called \textit{triangulation}.
The first step in the estimation of scene depth is finding correspondences between the two camera views, i.e., pixels in the two images that observe the same 3D position in the scene.
From these two corresponding points the 3D position of the point that gave rise to these two observations can be computed via triangulation, i.e., by intersecting two rays cast through the detected point correspondences.
Finding corresponding points between two different camera views is in general a highly challenging problem.
Normally, the search is based on local color descriptor matching or on solving an optimization problem.
One way to simplify this search is by exploiting the epipolar geometry between the two camera views.
This reduces the 2D search problem to a 1D search along a line.
Still, solving the correspondence problem requires sufficient local intensity and color variation in the recorded images, i.e., enough features.
Therefore, passive stereo reconstruction techniques work well in highly textured regions of the scene, but the search for correspondences might fail in featureless regions, which can result in missing depth information.
Active depth sensing approaches aim at alleviating this problem.

\index{Range Sensing!Active Sensing}
\section{Active Range Sensing} \label{MZsec:4}
Besides passive range sensing approaches, such as the stereo cameras discussed in the last section, there are also active techniques for range sensing.
Here, the term `active' refers to the fact that these sensors actively modify the scene to simplify the reconstruction problem.
There are two classes of active approaches \cite{active0}, which are based on different working principles, the so-called \textit{Time-of-Flight} (ToF) and \textit{Structured Light} (SL) cameras.
Structured Light cameras project a unique pattern into the scene to add additional features for matching and thus simplify feature matching and depth computation.
Therefore, they address the challenges passive reconstruction approaches face with featureless regions in the scene.
On the other hand, Time-of-Flight cameras emit a (potentially modulated) light pulse and measure its round trip time or phase shift.
Since Time-of-Flight cameras do not rely on color or texture to measure distance, they also do not struggle with texture-less scenes.
In both of the cases, modern commodity sensors normally work in the \textit{infrared} (IR) domain to not infer with human vision and enable the simultaneous capture of scene appearance.
In the following, we discuss both of these technologies in more detail and highlight their advantages and disadvantages.

\index{Range Sensing!Active Sensing!Time-of-Flight}
\index{Range Sensing!Active Sensing!Pulsed Time-of-Flight}
\index{Range Sensing!Active Sensing!Modulated Time-of-Flight}
\subsection{Time-of-Flight Sensors} \label{MZsec:5}
Besides passive binocular vision, many animals have implemented active range sensing approaches, e.g., the sonar used by whales is based on measuring the round-trip time of a sound wave.
As the name already suggests, the basic working principle of a Time-of-Flight camera is based on measuring the time of flight of an emitted light pulse \cite{tof0}.
More specifically, a light pulse is sent out from an emitter, it then traverses the scene until it hits an object and is reflected back to the Time-of-Flight camera, where a sensor records its arrival.
In general, there are two different types of Time-of-Flight cameras.

The first class, \textit{Pulsed Time-of-Flight cameras}, measures the round-trip time of a light pulse based on rapid shutters and a clock.
For Pulsed Time-of-Flight cameras, due to the constant known speed of light, the round trip distance can be computed by measuring the delay between sending and receiving the light pulse.
The scene depth can than be computed as half of the measured round trip distance:
\begin{equation}
	\text{Depth} = \frac{\text{Speed of Light} \times \text{Round Trip Time}}{2} \enspace{.}
\end{equation}
There are two types of pulsed Time-of-Flight cameras.
Point-wise Time-of-Flight sensors use a pan-tilt mechanism to obtain a time sequence of point measurements, this technique is also known as \textit{Light Detection And Ranging} (LiDAR).
Matrix-based Time-of-Flight cameras estimate a complete depth image for every time step based on a CMOS or CCD image sensor.
They employ light pulses generated by a laser that are a few nanoseconds apart.
Current commodity sensors belong to the second category, while Light Detection And Ranging is more employed for long range outdoor sensing, e.g., in the context of self driving cars.
Due to the immensely high speed of light of approximately 300,000 kilometers per second, the used clock for measuring the travel time has to be highly accurate, otherwise the depth measurements are imprecise.

The second type of Time-of-Flight camera uses a time-modulated light pulse and measures the phase shift between the emitted and returning pulse.
For \textit{Modulated Time-of-Flight} cameras, the light pulse is normally modulated by a continuous-wave.
A phase detector is used to estimate the phase of the returning light pulse.
Afterwards, the scene depth is obtained by the correlation between phase shift and scene depth.
Multi-frequency techniques can be employed to further improve the accuracy of the obtained depth measurements and the effective sensing range of the cameras.
Examples of current commodity Time-of-Flight cameras that are based on modulated time-of-flight include the Microsoft Kinect One and the Creative Senz3D.

\index{Range Sensing!Active Sensing!Structured Light}
\subsection{Structured Light Sensors} \label{MZsec:6}

Structured light sensing, similar to stereo reconstruction, is based on triangulation.
The key idea is to replace one of the two cameras in a stereo system by a projector.
The projector can be interpreted as an inverse camera.
By projecting a known unique structured pattern \cite{sl0} into the scene, additional artificial features are introduced into the scene.
This drastically simplifies correspondence matching, thus the quality of the reconstruction does not depend on the amount of natural color features in the scene.
Some sensors, such as the Microsoft Kinect, project a unique dot pattern \cite{sl1}, others project a temporal sequence of black and white stripes.
Structured Light cameras are wide spread and often used in research.
The commodity sensors of this category normally work in the infrared domain to not interfere with human vision and enable the simultaneous capture of an additional color image.
Examples of commodity sensors based on this technology are the Microsoft Kinect, Primesense Carmine, Asus Xtion Pro, and Intel Realsense.
Actually, the Intel Realsense is a hybrid of a passive and active sensing approach.
One problem of structured light cameras is that the sun's infrared radiation can saturate the sensor, making the pattern indiscernible.
This results in missing depth information.
The Intel Realsense alleviates this problem by combining active and passive vision.
To this end, it combines two infrared cameras with one infrared projector that is used to add additional features to the scene.
If the projector is overpowered by the ambient scene illumination the Intel Realsense defaults to standard stereo matching between two captured infrared images.
Normal working ranges for such commodity sensors are between 0.5\,meters to 12 meters. 
Similar to stereo systems, the accuracy of such sensors directly depends on the distance to the scene, i.e, the accuracy degrades with increasing distance.
The captured depth and color images of RGB-D sensors are not aligned, since the infrared and the color sensor are at different spatial locations, but the depth map can be mapped to the color image if the position and orientation of the two sensors is known.

\section{Comparison of the Sensing Technologies} \label{MZsec:7}
So far, we have discussed the most prevalent technologies for obtaining depth measurements.
More specifically, we had a look at passive stereo reconstruction and active structured light as well as time-of-flight sensing.
These three types of approaches are based on different physical and computational principles and thus have a different set of advantages and disadvantages.
For example, they have different working ranges and noise characteristics.
It is important to understand the advantages and disadvantages of the different technologies to be able to pick the right sensor for the application one wants to build.
In the following, we compare the discussed three technologies in detail.

\subsection{Passive Stereo Sensing}

Stereo reconstruction is based on finding correspondences between points observed in both camera views and triangulation to obtain the depth measurements.
Thus, the quality and density of the depth map directly depends on the amount of color and texture features in the scene.
More specifically, the quality and density of the depth measurements degrades with a decreasing amount of available features.
One extreme case, that is often found in indoor scenes, are walls of uniform color, which can not be reconstructed, since no reliable matches between the left and right camera can be found. 
Similar to uniformly colored objects, also low light, e.g., scanning in a dark room, can heavily impact the ability to compute reliable matches.
Repeated structures and symmetries in the scene can lead to wrong feature associations.
In this case, multiple equally good matches exist and sophisticated pruning strategies and local smoothness assumptions are required to select the correct match.
Passive stereo is a triangulation-based technique.
Therefore, it requires a baseline between the two cameras, which leads to a larger form factor of the device.
Similar to all approaches based on triangulation, the quality of the depth measurements degrades with increasing distance to the scene and improves for larger baselines.
The noise characteristics of stereo vision systems have been extensively studied \cite{stereonoise0}.
One significant advantage of passive stereo systems is that multiple devices do not interfere with each other.
This is in contrast to most active sensing technologies.
In addition, stereo sensing can have a large working range if a sufficiently large baseline between the two cameras is used.
Since stereo systems are built from off-the-shelf monochrome or color cameras, they are cheap to build and are quite energy efficient.

\subsection{Structured Light Sensing}

Active range sensing techniques, such as structured light sensing, remove one of the fundamental problems of passive approaches, i.e., the assumption that the scene naturally contains a large amount of color or texture features.
This is made possible, since the projected pattern introduces additional features into the scene which can be used for feature matching.
For example, this allows to reconstruct even completely uniformly colored objects, but comes at the price of a higher energy consumption of the sensor, since the scene has to be actively illuminated.
In addition, structured light sensors do not work under strong sunlight, since the sensor will be oversaturated by the sun's strong IR radiation and thus the projected pattern is not visible.
Due to the projection of a structured pattern, a few problems might occur:
If the projected pattern is partially occluded from the sensor's viewpoint, which is especially a problem at depth discontinuities in the scene, the depth cannot be reliably computed.
Normally, this leads to missing depth estimates around the object silhouette, which leads to a slightly `shrunken' reconstruction.
This also complicates the reconstruction of thin objects.
The projected pattern might also be absorbed by dark objects, reflected by specular objects, or refracted by transparent objects, all of these situations might lead to wrong or missing depth estimates.
Active structured light depth sensing technology has a limited working range, normally up to 15 meters, since otherwise too much energy would be required to consistently illuminate the scene.
The noise characteristics of structured light sensors have been extensively studied \cite{kinectnoise0,kinectnoise1}.
Using multiple sensors at the same time might result in a loss of depth accuracy due to interference of multiple overlapping patterns, since the correspondences can not be reliably computed.
Geometric structures that are smaller than the distance between the projected points are lost.


\subsection{Time-of-Flight Sensing}

In contrast to stereo vision and structured light, Time-of-Flight cameras are based on a different physical measurement principle, i.e., measuring time-of-flight/phase-shift of a light pulse instead of triangulation.
This leads to a different set of failure modes and drastically different noise characteristics.
One of the biggest artifacts in time-of-flight depth images are the so called `flying pixels' at depth discontinuities.
Flying pixels have depth values between the fore- and background values that exist in reality.
They appear if the light pulse is reflected back by multiple parts of the scene and then measured at the same sensor location.
This is related to the much wider class of multi-path interference effects ToF cameras suffer from, i.e, multiple indirect light paths being captured by the sensor.
Examples of this are multi-path effects caused by materials that exhibit reflections or refractions, e.g., mirrors or glass.
Even in relatively diffuse scenes, indirect bounces of the light pulse might influence the reconstruction quality.
Dark materials do not reflect light.
Therefore, no returning light pulse can be measured which leads to holes in the depth map.
Similar to other active sensing modalities, Time-of-Flight suffers from interference between multiple sensors if they use the same phase-shift.
This can be alleviated by using different modulation frequencies for each sensor.
Similar to active Structured Light, Time-of-Flight depth sensing struggles under strong sunlight.
Since Time-of-Flight cameras require a certain integration time to obtain a good signal-to-noise ratio, fast motions lead to motion-blurred depth estimates.
The noise characteristics of Time-of-Flight cameras have been extensively studied \cite{tofnoise0}.


\bibliographystyle{spmpsci}

\bibliography{Zollhoefer}

\end{document}